 \documentclass[smallabstract,smallcaptions]{dccpaper}

\usepackage{url}            
\usepackage{booktabs}       
\usepackage{amsfonts}       
\usepackage{nicefrac}       
\usepackage{microtype}      
\usepackage{xcolor}         
\usepackage{enumitem}
\usepackage{wrapfig}
\usepackage{subfigure}
\usepackage{algorithmic}
\usepackage{graphicx}
\usepackage{amsmath,bm}
\usepackage{multirow}
\usepackage{multicol}

\newlength{\figurewidth}
\newlength{\smallfigurewidth}

\begin{document}

\title
{\large
\textbf{Learning to Compress Unmanned Aerial Vehicle (UAV) Captured Video: Benchmark and Analysis}
}

\author{%
Chuanmin Jia$^{\dag}$, Feng Ye$^{\dag}$, Huifang Sun$^{\ddagger}$, Siwei Ma$^{\dag,\ddagger}$, Wen Gao$^{\dag,\ddagger}$\\[0.5em]
{\small\begin{minipage}{\linewidth}\begin{center}
\begin{tabular}{ccc}
$^{\dag}$Peking University, Beijing, China, $^{\ddagger}$Peng Cheng Laboratory, Shenzhen, China\\ 
\small
\end{tabular}
\end{center}\end{minipage}}
}

\maketitle
\thispagestyle{empty}

\begin{abstract}
During the past decade, the Unmanned-Aerial-Vehicles (UAVs) have attracted increasing attention due to their flexible, extensive, and dynamic space-sensing capabilities. The volume of video captured by UAVs is exponentially growing along with the increased bitrate generated by the advancement of the sensors mounted on UAVs, bringing new challenges for on-device UAV storage and air-ground data transmission. Most existing video compression schemes were designed for natural scenes without consideration of specific texture and view characteristics of UAV videos. In this work, we first contribute a detailed analysis of the current state of the field of UAV video coding. Then we propose to establish a novel task for learned UAV video coding and construct a comprehensive and systematic benchmark for such a task, present a thorough review of high quality UAV video datasets and benchmarks, and contribute extensive rate-distortion efficiency comparison of learned and conventional codecs after. Finally, we discuss the challenges of encoding UAV videos. It is expected that the benchmark will accelerate the research and development in video coding on drone platforms.

\end{abstract}

\Section{Introduction}
Drones (or Unmanned Aerial Vehicles, UAVs) are receiving increasing attention during the past decade, due to their flexible, extensive, and dynamic space sensing availability. The commercial UAV market report describes that the global commercial drone market size will reach 501.4 billion by 2028~\cite{zhu2021detection}. These drones can be efficiently used for autonomous facility maintenance, scientific discoveries, or agricultural monitoring via the equipped cameras. Consequently, high efficiency compression and low-cost storage scheme of visual data collected from drones become highly demanding, which brings video coding to UVAs even more closely. 

Lossy video compression using neural networks has been popular research topic in recent years, the objective of which is to achieve the compact representation with minimal bit-rate (entropy) while preserving maximal video signal quality, yielding the typical rate-distortion (R-D) optimization problem. The video compression community has continued to gain coding performance along with the local module advancement of hybrid coding framework in a standardized fashion~\cite{wiegand2003overview,sullivan2012overview,ma2015avs2,zhang2019recent,chen2018overview}. Recently, with the tremendous success in artificial intelligence, neural video compression (NVC) attracts increasing attention from both academic and industry realms for the capability of end-to-end global optimization. 

However, it is observed that fewer efforts have been dedicated to the field of UAV video coding, leaving such research topic less studied. Existing UAV-oriented video coding methods tend to build auxiliary modules or particular-optimized algorithms on the top of existing video coding standards and their extensions~\cite{belyaev2019efficient,bhaskaranand2014global} to adapt the specific content characteristics of UAV videos. The motion model capacity between learned codecs and conventional codecs may be quite different. Indeed, the pixel-wise fine-grained motion is more flexible for drone video compression to describe aerial motion compared with the commonly used hybrid coding techniques like block-based motion-compensation prediction., the pixel-wise fine-grained motion is more flexible for the drone video compression to describe aerial motion. Moreover, considering that the UAV videos have completely different field-of-view, motion dynamics, and viewpoint distance against compared to natural videos, the compression of variable-size block-wise motion representation in most existing hybrid coding approaches will consume a large number of bits for compression. Therefore, it is nontrivial to develop a new framework for UAV videos.

In this paper, we propose to build a novel task named learning based UAV video coding. Furthermore, we extensively collect the UAV videos with different content variations, including in-door and out-door scenes, object-scale variations and viewpoint distance, different climate condition etc. Then we encode those properly-selected videos using popular end-to-end optimized video codecs and conventional hybrid codecs, to form a comprehensive benchmark for learned drone video compression. We also provide a detailed analysis and envision the challenge of such task for future research. The main contributions of this paper are three folds. {\bf First}, we construct a comprehensive benchmark for the task of drone video compression which consist of the rate-distortion (R-D) behavior of both hybrid and learned video codecs. To our knowledge, it is the first attempt in end-to-end optimized solution to compress drone videos. 
{\bf Second}, we provide the review and analysis of the learned drone video compression schemes and further discuss the challenges of encoding UAV videos. {\bf Third}, this benchmark and related research is accomplished as a milestone MPAI\footnote{https://mpai.community/} End-to-end Video (EEV) coding project. The proposed benchmark has constructed a solid baseline for compressing UAV videos and facilitates the future research works for related task.

\Section{Learned UAV Video Coding Benchmark}

\begin{figure}[]
    \centering
	\includegraphics[width=0.9\textwidth]{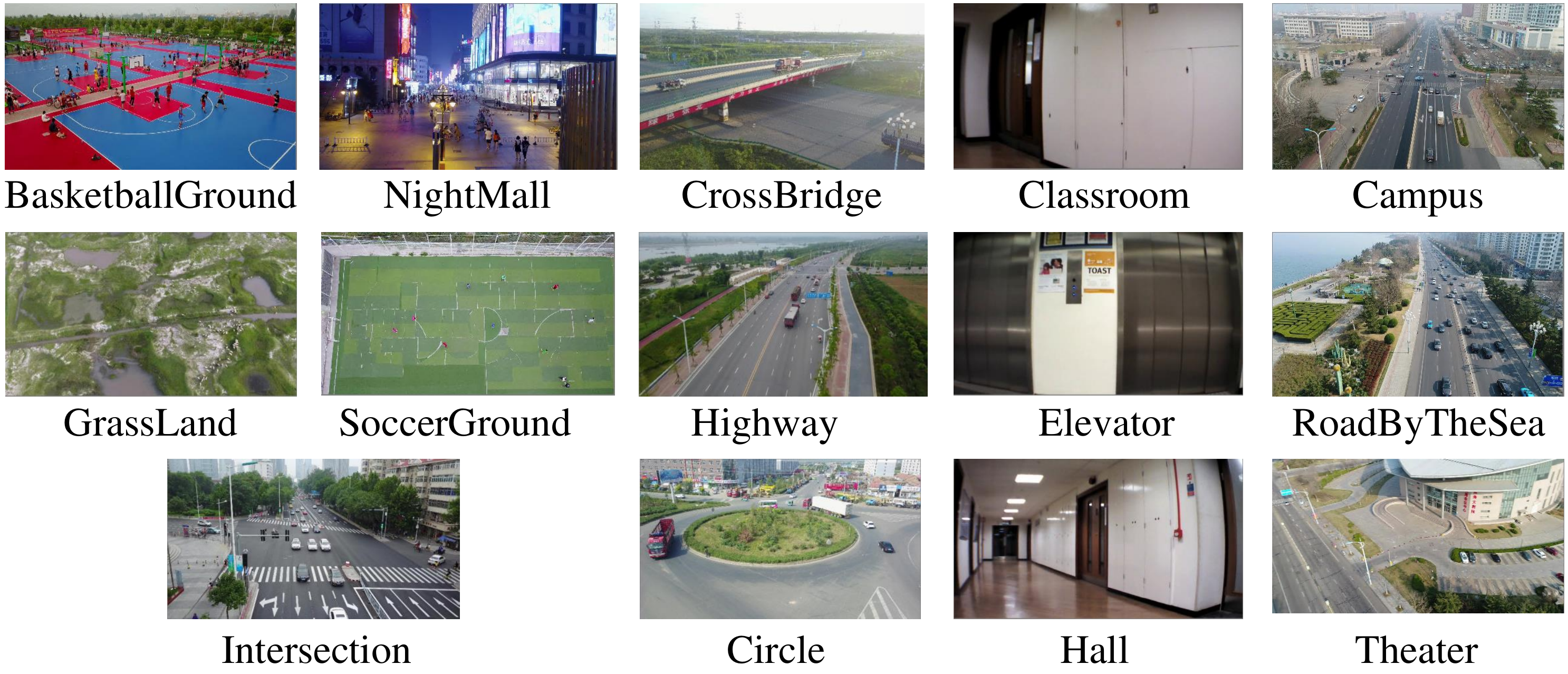}
	\caption{The snapshot of the UAV sequences with different content characteristics.}
	\label{thumbnail}
\end{figure}

{\bf Video Sequences Characteristics.} Prior UAV related research mainly focuses on high level computer vision tasks for on-device cameras, such as single/multiple object detection, image/video object tracking, etc. Notably, there are plenty of workshops and grand challenges that aim to provide large-scale drone video datasets, with thousands of images frames and still images. Given those high quality UAV captured data, we collect a set of video sequences to build the UAV video coding benchmark from those diverse contents, considering the record device type (various model of drone-mounted cameras), diverse in many aspects including location (in-door and out-door places), environment (traffic workload, urban and rural regions), objects (e.g., pedestrian and vehicles), and scene object density (sparse and crowded scenes). For better understanding of those videos, we provide comprehensive summary of the prepared learned drone video coding benchmark in Table~\ref{tab:seqs}. The corresponding thumbnail of each clip is depicted in Fig.~\ref{thumbnail} as supplementary information. There are 14 video clips from multiple different source UAV video datasets~\cite{zhu2021detection,kouris2019informed,du2018unmanned}. The resolutions and frame rate of them range from $2720\times1520$ down to $640\times352$, 24 to 30 respectively. The frame count of each test video is 100. 

\begin{table}[]
\caption{Video sequence characteristics of the proposed learned UAV video coding benchmark}
\centering
\footnotesize
\begin{tabular}{|c|c|c|c|c|c|c|}
\hline
\textbf{Source}                                                                   & \textbf{\begin{tabular}[c]{@{}c@{}}Sequence\\ Name\end{tabular}} & \textbf{\begin{tabular}[c]{@{}c@{}}Spatial \\ Resolution\end{tabular}} & \textbf{\begin{tabular}[c]{@{}c@{}}Frame \\ Count\end{tabular}} & \textbf{\begin{tabular}[c]{@{}c@{}}Frame \\ Rate\end{tabular}} & \textbf{\begin{tabular}[c]{@{}c@{}}Bit\\ Depth\end{tabular}} & \textbf{\begin{tabular}[c]{@{}c@{}}Scene\\ Feature\end{tabular}} \\ \hline
\multirow{5}{*}{\begin{tabular}[c]{@{}c@{}}Class A\\VisDrone-SOT\\ TPAMI2021~\cite{zhu2021detection}\end{tabular}} & BasketballGround                                                 & 960x528                                                                & 100                                                             & 24                                                             & 8                                                            & Outdoor                                                          \\ \cline{2-7} 
                                                                                  & GrassLand                                                        & 1344x752                                                               & 100                                                             & 24                                                             & 8                                                            & Outdoor                                                          \\ \cline{2-7} 
                                                                                  & Intersection                                                     & 1360x752                                                               & 100                                                             & 24                                                             & 8                                                            & Outdoor                                                          \\ \cline{2-7} 
                                                                                  & NightMall                                                        & 1920x1072                                                              & 100                                                             & 30                                                             & 8                                                            & Outdoor                                                          \\ \cline{2-7} 
                                                                                  & SoccerGround                                                     & 1904x1056                                                              & 100                                                             & 30                                                             & 8                                                            & Outdoor                                                          \\ \hline
\multirow{3}{*}{\begin{tabular}[c]{@{}c@{}}Class B\\VisDrone-MOT\\ TPAMI2021~\cite{zhu2021detection}\end{tabular}} & Circle                                                           & 1360x752                                                               & 100                                                             & 24                                                             & 8                                                            & Outdoor                                                          \\ \cline{2-7} 
                                                                                  & CrossBridge                                                      & 2720x1520                                                              & 100                                                             & 30                                                             & 8                                                            & Outdoor                                                          \\ \cline{2-7} 
                                                                                  & Highway                                                          & 1344x752                                                               & 100                                                             & 24                                                             & 8                                                            & Outdoor                                                          \\ \hline
\multirow{3}{*}{\begin{tabular}[c]{@{}c@{}}Class C\\Corridor\\ IROS2018~\cite{kouris2019informed}\end{tabular}}      & Classroom                                                        & 640x352                                                                & 100                                                             & 24                                                             & 8                                                            & Indoor                                                           \\ \cline{2-7} 
                                                                                  & Elevator                                                         & 640x352                                                                & 100                                                             & 24                                                             & 8                                                            & Indoor                                                           \\ \cline{2-7} 
                                                                                  & Hall                                                             & 640x352                                                                & 100                                                             & 24                                                             & 8                                                            & Indoor                                                           \\ \hline
\multirow{3}{*}{\begin{tabular}[c]{@{}c@{}}Class D\\UAVDT\_S\\ ECCV2018~\cite{du2018unmanned}\end{tabular}}      & Campus                                                           & 1024x528                                                               & 100                                                             & 24                                                             & 8                                                            & Outdoor                                                          \\ \cline{2-7} 
                                                                                  & RoadByTheSea                                                     & 1024x528                                                               & 100                                                             & 24                                                             & 8                                                            & Outdoor                                                          \\ \cline{2-7} 
                                                                                  & Theater                                                          & 1024x528                                                               & 100                                                             & 24                                                             & 8                                                            & Outdoor                                                          \\ \hline
\end{tabular}
\label{tab:seqs}
\end{table}

The dataset contains various UAV videos captured under different conditions, including environments, flight altitudes, and camera views, e.g., basketball ground, highway, classroom and campus. These video clips are selected from several categories of real-life objects in different scene object densities and lighting conditions, representing diverse scenarios in our daily life. Compared to natural videos, UAV-captured videos are generally recorded by drone-mounted cameras in motion and at different viewpoints and altitudes. These features bring several new challenges, such as motion blur, scale changes and complex background. Heavy occlusion, non-rigid deformation and tiny scales of objects might be of great challenge to 
drone video compression.

Different from natural videos, the UAV videos is not viewpoint and scale invariant. Thus the predictive coding efficiency might be restricted due to large or global motion. Moreover, the distortion characteristic (bird/fish eye view) of the drone-equipped camera also makes UAV video distinct from natural ones. Therefore, conventional hybrid video coding with block-based motion compensation might not be ideal choice when compressing them. It is observed from Fig.~\ref{thumbnail} that the in-door drone captured videos are highly distorted around the view boundary and the deformation of objects also degrades the inter-prediction efficacy, resulting more intra coded blocks after rate-distortion optimization based mode decision during encoding procedure.

{\bf Pre-processing.} The drone videos usually have slight compression effect by the on-device hardware codec. To overcome this issue and produce near pristine video clips, we conduct a set of pre-processing procedures to those drone videos including down-sampling, boundary cropping, and color space alignment (in RGB domain). We further particular center-cropped the resolutions to be multiple of 16 to be compatible with convolution operations in learned codecs. Eventually we employ the RGB planar data structure to store the raw data. It should also be pointed out that the processing color space is RGB in this benchmark, including encoding, decoding, and quality assessment. 

{\bf Codecs selection.} To comprehensively reveal R-D efficiency of UAV video using both conventional and learned codecs, we encode the above collected drone video sequences using HEVC/H.265 reference software with screen content coding (HEVC-SCC) extension reference software (HM-16.20-SCM-8.8)~\cite{xu2015overview} and the emerging learned video coding framework OpenDVC~\cite{lu2019dvc,yang2020opendvc}. Moreover, the reference model of MPAI End-to-end Video (EEV) is also employed to compress the UAV videos. As such, the baseline coding results are composed of three different codecs. 

\begin{figure}[]
    \centering
	\subfigure[]{
	\includegraphics[width=0.32\textwidth]{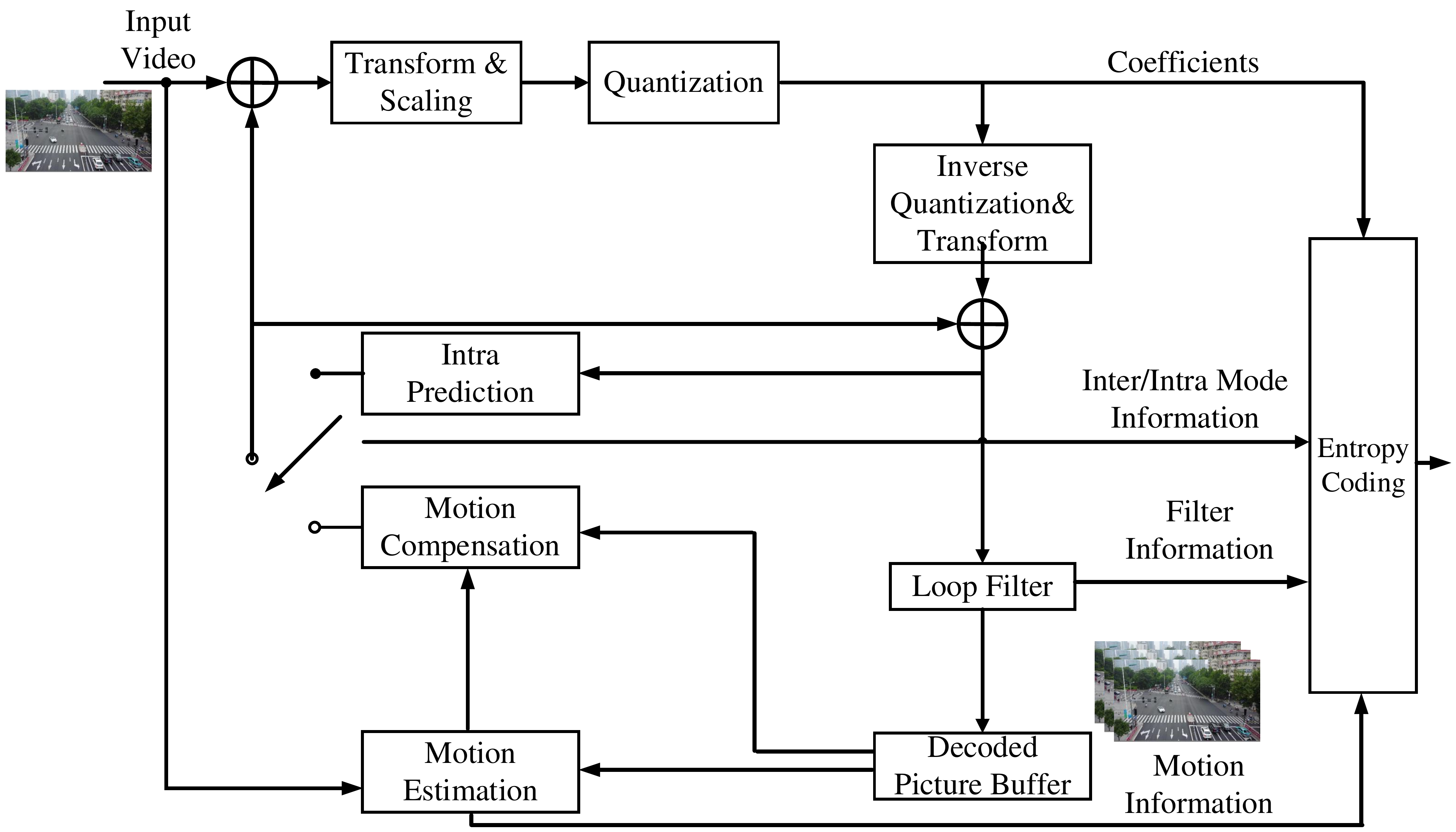}}
	\subfigure[]{
	\includegraphics[width=0.32\textwidth]{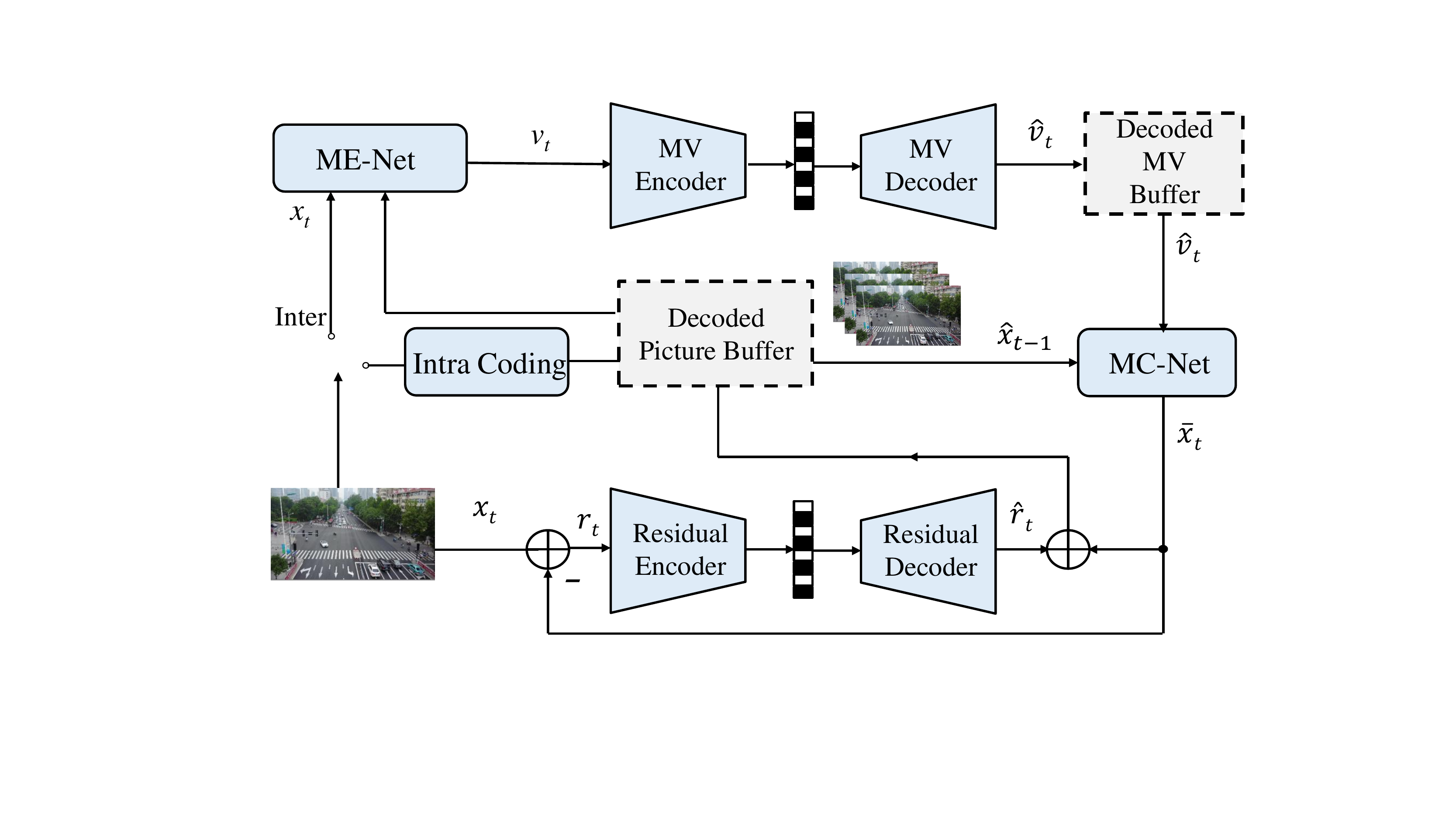}}	
	\subfigure[]{
	\includegraphics[width=0.26\textwidth]{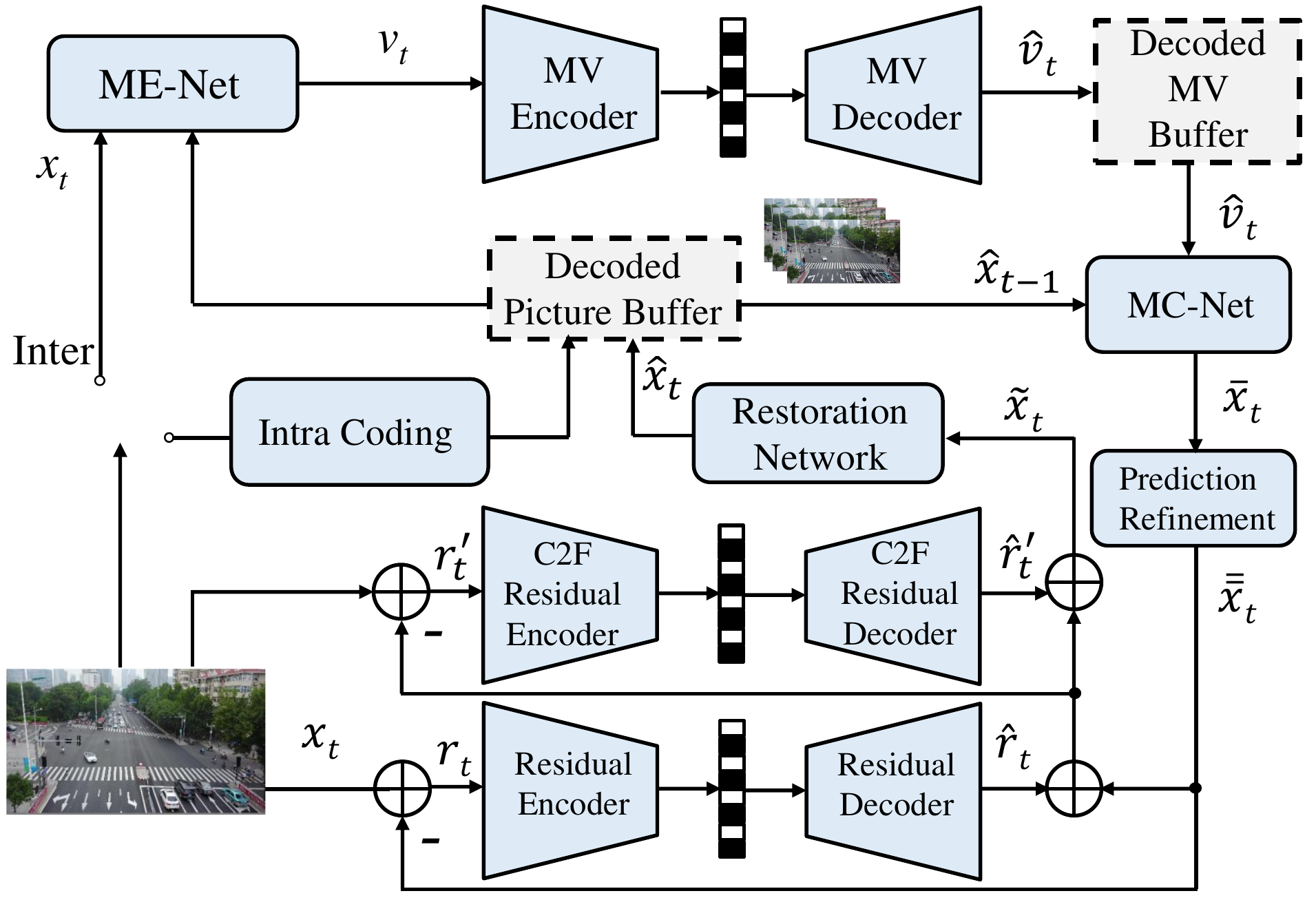}}		
	\caption{Block diagram of different codecs. (a) Conventional hybrid codec HEVC. (b) OpenDVC. (3) MPAI EEV. Zoom-in for better visualization.}
	\label{codecs}
\end{figure}

The schematic diagrams of them are shown in Fig.~\ref{codecs}. The left panel represents the classical hybrid codec. The remaining two are learned codecs, OpenDVC and EEV respectively. It is easy to observe that the EEV software is an enhanced version of OpenDVC codec by incorporating more advanced modules such as motion compensation prediction improvement, two-stage residual modeling, and in-loop restoration network. Specifically, ${x_{t}}$ and $\hat{x}_{t}$ denote the un-coded and restored image in both Fig.~\ref{codecs}(b) and Fig.~\ref{codecs}(c) while $\tilde{x}_{t}$ is the predicted frame. Analogously, ${v_{t}}$ and $\hat{v}_{t}$ correspond to the pristine and restored optical flow. ${r_{t}}$ and $\hat{r}_{t}$
are the uncompressed and reconstructed residual. Moreover, ${{r}^{'}_{t}}$ and $\hat{r}^{'}_{t}$ are coarse-to-fine (C2F) residual proposed by EEV. Therefore, the reconstruction of learned codecs for OpenDVC and EEV can be represented as $\hat{x}_{t}$=$\tilde{x}_{t}$+$\hat{r}_{t}$ and $\hat{x}_{t}$=$\tilde{x}_{t}$+$\hat{r}_{t}$+$\hat{r}^{'}_{t}$ respectively.

Another important factor for learned codecs is train-and-test data consistency. It is widely accepted in machine learning community that train and test data should be independent identically distributed. However, both OpenDVC and EEV are trained using natural video dataset vimeo-90k~\cite{xue2019video} with mean-square-error (MSE) as distortion metrics. We employ those pre-trained weights of learned codecs without fine-tuning them on drone video data to guarantee the generalisation of the benchmark. By doing so, the effectiveness and drawbacks of learned and conventional codecs could be directly observed when comparing the coding performances.

{\bf Parameter Setting.} We follow the common test condition (CTC) to utilize lowdelay-P configuration for HEVC codec. The main-RExt profile is utilized. We set the $InputColourSpaceConvert$ parameter as $RGBtoGBR$ when encoding. When decoding, we set the $OutputInternalColourSpace$ as $GBRtoRGB$ to ensure the consistency of reconstruction results and decoded results. The internal bit-depth is set to be 8 because the drone-captured videos are all using standard-dynamic-range (SDR). The quantization parameters (QPs) are set to be \{30,34,38,42\} to realize four different bit-rate points. Regarding the OpenDVC and EEV software, the variable-rate coding is realized by training multiple models using the hyper-parameter $\lambda$=\{2048,1024,512,256\}, whose functionality is to balance the trade-off between coding bits and distortion. It is worth noting that the intra period (IP) number is set to be 16 for all codecs for reasonable comparison. Note that when evaluating learned codecs, the $mode$ and $metric$ parameter are set to be \textit{PSNR}. $seq\_wid$ and $seq\_hgt$ indicate the resolution of each sequence. In summary, the command line settings for HM-16.20-SCC-8.8, OpenDVC, and EEV are listed below.

\begin{itemize}
\item TAppEncoder -c encoder\_LDP.cfg -InputBitDepth 8
 -InputChromaFormat 444 -Level 6.2 -wdt seq\_wid -hgt seq\_hgt -f 100 -fr fps -q QP -IntraPeriod 16 -InputColourSpaceConvert RGBtoGBR -SNRInternalColourSpace 1\\ -OutputColourSpaceConvert GBRtoRGB
\item python test\_opendvc.py -path seqname -mode PSNR -IntraPeriod 16 -metric PSNR -l $\lambda$
\item python test\_eev.py -path seqname -mode PSNR -IntraPeriod 16 -metric PSNR -l $\lambda$
\end{itemize}

{\bf Evaluation Method.} Since all drone videos in our proposed benchmark are using RGB color space, the quality assessment methods are also applied to the reconstruction in RGB domain. For each frame, the peak-signal-noise-ratio (PSNR) 
values of $x_{t}$ and $\hat{x}_{t}$ are calculated for each component channel then the RGB averaged value is obtained to indicate its picture quality. Regarding bitrate, we calculate bit-per-pixel (BPP) using the binary files produced by codecs.
We report the coding efficiency of different codecs using the Bjøntegaard delta bit rate (BD-rate) measurement~\cite{bjontegaard2001calculation}.

\Section{R-D Results and Performance Analysis}
\SubSection{Coding Performance}
The corresponding PSNR based R-D performances of the three different codecs are shown in Table~\ref{tab:rdperformance}. Regarding the simulation results, it is observed that around 20\% bit-rate reduction could be achieved when comparing EEV and OpenDVC codec. This shows promising performances for the learned codecs and its improvement made by EEV software. By comparatively studying the coding gain of different UAV videos, it is observed that the R-D efficiency is consistent among diverse kinds of contents. This point corresponds to our previous assumption that learning based codecs have partial generalisation ability even without fine-tuning on drone video datasets. Some minor turbulence is perceived in Class C, which confirms that in-door UAV content has different texture characteristics with out-door ones.

\begin{table}[]
\centering
\footnotesize
\caption{The BD-rate performance of different codecs (OpenDVC, EEV, and HM-16.20-SCM-8.8) on drone video compression. The distortion metric is RGB-PSNR.}
\begin{tabular}{|cc|c|c|}
\hline
\multicolumn{1}{|c|}{\textbf{Category}}                                                               & \textbf{\begin{tabular}[c]{@{}c@{}}Sequence\\ Name\end{tabular}} & \textbf{\begin{tabular}[c]{@{}c@{}}BD-Rate Reduction\\ EEV vs OpenDVC\end{tabular}} & \textbf{\begin{tabular}[c]{@{}c@{}}BD-Rate Reduction\\ EEV vs HEVC\end{tabular}} \\ \hline
\multicolumn{1}{|c|}{\multirow{5}{*}{\begin{tabular}[c]{@{}c@{}}Class A\\ VisDrone-SOT\end{tabular}}} & BasketballGround                                                 & -23.84\%                                                                            & 9.57\%                                                                           \\ \cline{2-4} 
\multicolumn{1}{|c|}{}                                                                                & GrassLand                                                        & -16.42\%                                                                            & -38.64\%                                                                         \\ \cline{2-4} 
\multicolumn{1}{|c|}{}                                                                                & Intersection                                                     & -18.62\%                                                                            & -28.52\%                                                                         \\ \cline{2-4} 
\multicolumn{1}{|c|}{}                                                                                & NightMall                                                        & -21.94\%                                                                            & -6.51\%                                                                          \\ \cline{2-4} 
\multicolumn{1}{|c|}{}                                                                                & SoccerGround                                                     & -21.61\%                                                                            & -10.76\%                                                                         \\ \hline
\multicolumn{1}{|c|}{\multirow{3}{*}{\begin{tabular}[c]{@{}c@{}}Class B\\ VisDrone-MOT\end{tabular}}} & Circle                                                           & -20.17\%                                                                            & -25.67\%                                                                         \\ \cline{2-4} 
\multicolumn{1}{|c|}{}                                                                                & CrossBridge                                                      & -23.96\%                                                                            & 26.66\%                                                                          \\ \cline{2-4} 
\multicolumn{1}{|c|}{}                                                                                & Highway                                                          & -20.30\%                                                                            & -12.57\%                                                                         \\ \hline
\multicolumn{1}{|c|}{\multirow{3}{*}{\begin{tabular}[c]{@{}c@{}}Class C\\ Corridor\end{tabular}}}     & Classroom                                                        & -8.39\%                                                                             & 178.49\%                                                                         \\ \cline{2-4} 
\multicolumn{1}{|c|}{}                                                                                & Elevator                                                         & -19.47\%                                                                            & 109.54\%                                                                         \\ \cline{2-4} 
\multicolumn{1}{|c|}{}                                                                                & Hall                                                             & -15.37\%                                                                            & 58.66\%                                                                          \\ \hline
\multicolumn{1}{|c|}{\multirow{3}{*}{\begin{tabular}[c]{@{}c@{}}Class D\\ UAVDT\_S\end{tabular}}}     & Campus                                                           & -26.94\%                                                                            & -25.68\%                                                                         \\ \cline{2-4} 
\multicolumn{1}{|c|}{}                                                                                & RoadByTheSea                                                     & -20.98\%                                                                            & -24.40\%                                                                         \\ \cline{2-4} 
\multicolumn{1}{|c|}{}                                                                                & Theater                                                          & -19.79\%                                                                            & 2.98\%                                                                           \\ \hline
\multicolumn{2}{|c|}{\textbf{Class A}}                                                                                                                                   & \textbf{-20.49\%}                                                                   & \textbf{-14.97\%}                                                                \\ \hline
\multicolumn{2}{|c|}{\textbf{Class B}}                                                                                                                                   & \textbf{-21.48\%}                                                                   & \textbf{-3.86\%}                                                                 \\ \hline
\multicolumn{2}{|c|}{\textbf{Class C}}                                                                                                                                   & \textbf{-14.41\%}                                                                   & \textbf{115.56\%}                                                                \\ \hline
\multicolumn{2}{|c|}{\textbf{Class D}}                                                                                                                                   & \textbf{-22.57\%}                                                                   & \textbf{-15.70\%}                                                                 \\ \hline
\multicolumn{2}{|c|}{\textbf{Average}}                                                                                                                                   & \textbf{-19.84\%}                                                                   & \textbf{15.23\%}                                                                 \\ \hline
\end{tabular}
\label{tab:rdperformance}
\end{table}

When we directly compare the coding performance of EEV and HEVC, obvious performance gap between the in-door and out-door sequences could be observed. Generally speaking, the HEVC SCC codec outperforms the learned codec by 15.23\% over all videos. Regarding Class C, EEV is significantly inferior to HEVC by clear margin, especially for the $Classroom$ and $elevator$ sequences. The reason for such results might be domain shift of the train-test data, such as content difference (natural scene for training and in-door drone video for test) and imaging type change (in-door drone videos are highly distorted around the boundary). However, the EEV codec has better coding efficiency than HEVC in Class A, Class B and Class D by 14.97\%, 3.86\% and 15.70\% respectively, which shows great potentials for the learning based UAV video codec. It is additionally observed that the coding performances are not consistency in different drone sequences within a certain class. EEV outperforms HEVC by over 38\% in $GrassLand$ but less performed in $BasketballGround$ by 9.5\%. Such R-D statistics reveal that learned codecs are more sensitive to the video content variations than conventional hybrid codecs if we directly apply natural-video-trained codec to UAV video coding. For future research, this point could be resolved and modeled as an out-of-distribution problem and extensive knowledge could be borrowed from the machine learning community.

\begin{figure}[!htb]
    \centering
	\subfigure[Campus 1024$\times$528]{
	\includegraphics[width=0.4\textwidth]{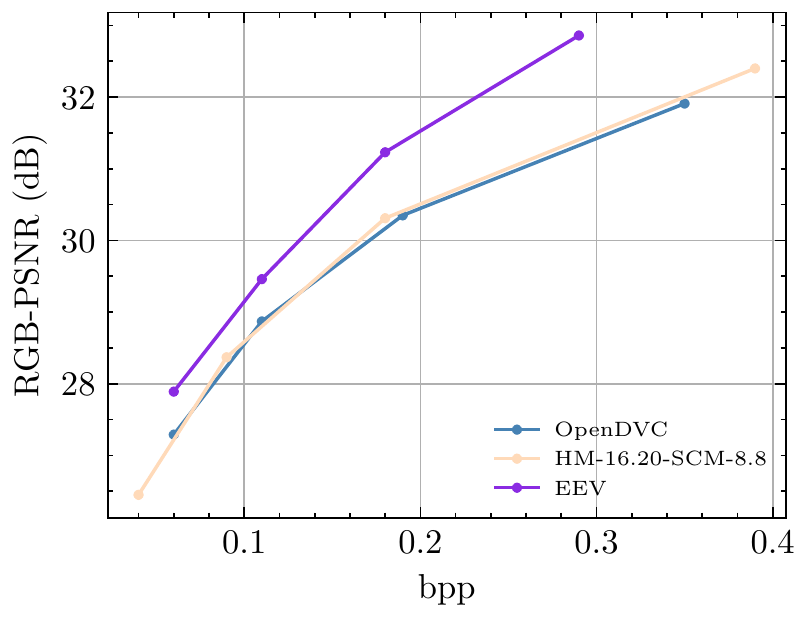}}
	\subfigure[Highway 1344$\times$752]{
	\includegraphics[width=0.395\textwidth]{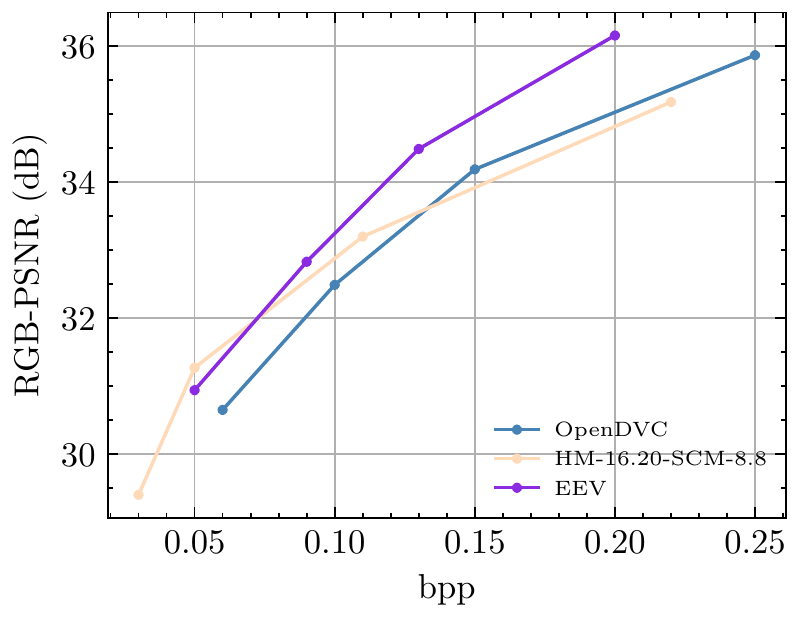}}
	\subfigure[NightMall 1920$\times$1072]{
	\includegraphics[width=0.4\textwidth]{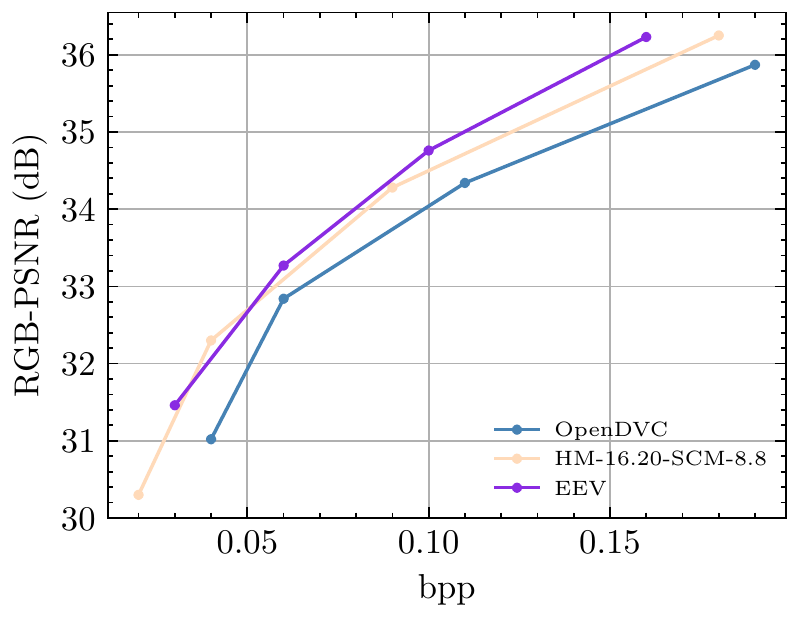}}
	\subfigure[Elevator 640$\times$352]{
	\includegraphics[width=0.42\textwidth]{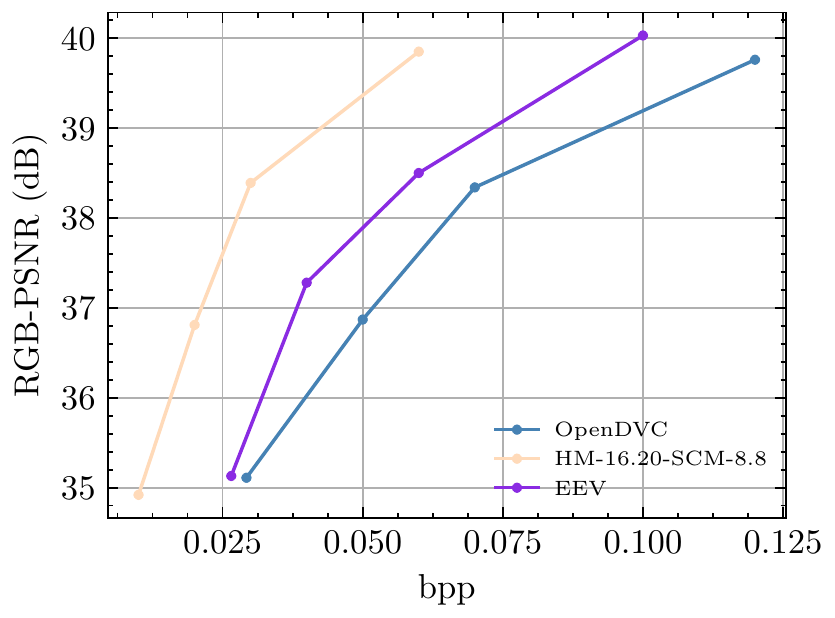}}	
	\subfigure[SoccerGround 1904$\times$1056]{
	\includegraphics[width=0.4\textwidth]{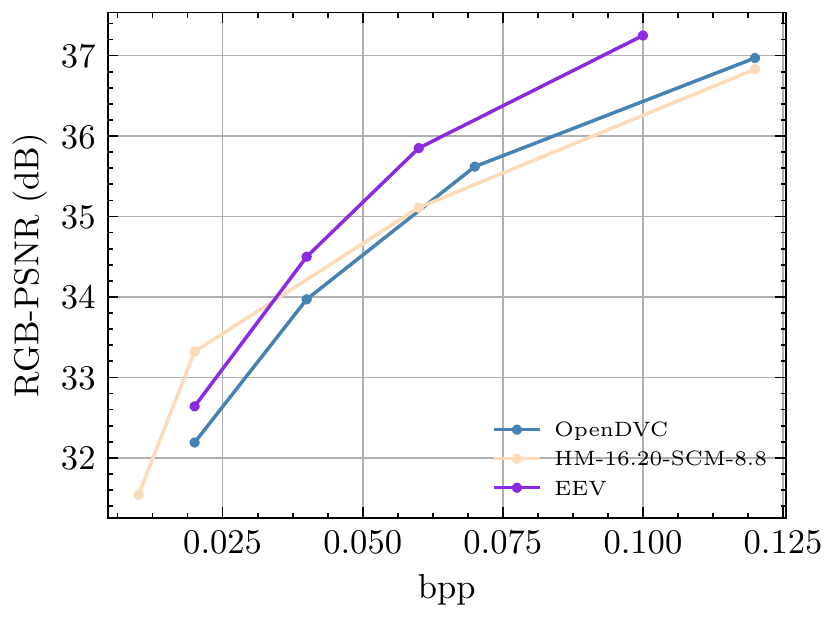}}
	\subfigure[Intersection 1360$\times$752]{
	\includegraphics[width=0.38\textwidth]{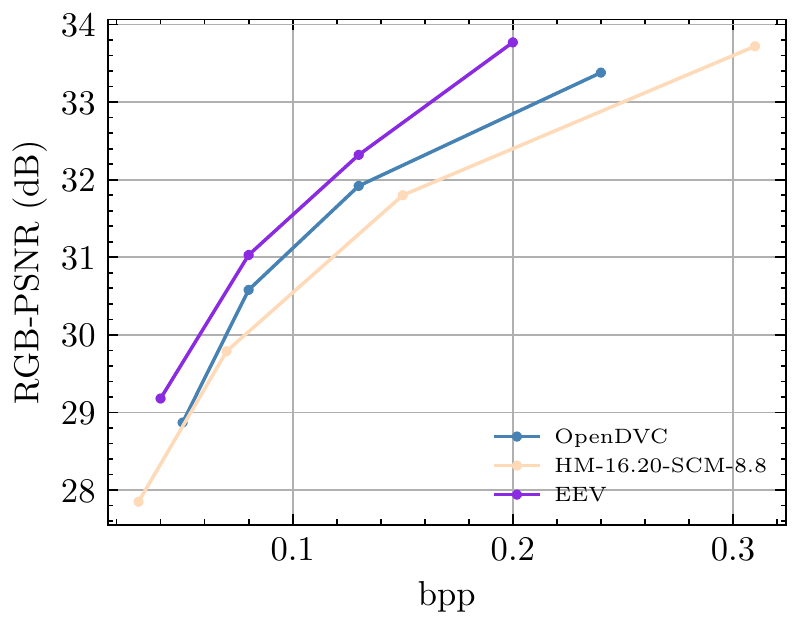}}
	
	\caption{R-D curves of different codecs (OpenDVC~\cite{yang2020opendvc}, EEV, and HM-16.20-SCM-8.8~\cite{xu2015overview}) for difference drone video sequences. The blue-violet, peach-puff, and steel-blue curves denote EEV, HEVC and OpenDVC codec respectively.}
	\label{fig:rdcurves}
\end{figure}

\SubSection{R-D Curves and Analysis}
To further dive into the R-D efficiency interpretation of different codecs, we plot the R-D curves of different methods in Fig.~\ref{fig:rdcurves}. Specifically, we select one test video clip from each class to illustrate the diversity and R-D behavior. The R-D curves of $Camplus$, $Highway$, $NightMall$, $Elevator$, $SoccerGround$, and $Intersection$ are depicted. The blue-violet, peach-puff, and steel-blue curves denote EEV, HEVC and OpenDVC codec respectively. Given the BD-rate performances and R-D curves, interesting observations and phenomenon could be perceived. It is observed from those curves that the EEV consistently outperforms the OpenDVC for all bit-rate points and all sequences. However, when comparing with HM-16.20-SCM-8.8 codec, the EEV codec might be less effective is some cases (especially for bpp less than 0.1 scenarios). Regarding the in-door test video, the conventional codec performs much better than the remaining two learned codecs. As aforementioned, the videos capture by in-door-drone-equipped cameras are highly distorted at boundary area due to fish-eye imaging effect. The learned codecs are optimized using natural videos, resulting domain gap. Therefore, the content characteristic of UAV videos and its distance to the natural videos shall be modeled and investigated in future research.

Classical codecs are designed to be content independent by using crafted rules and non-parametric models, such that different contents are treated equal during deployment, which guarantees the robustness of them. For data-trained codecs, the generalisation ability relies on both of the model-designation principles and the training data as well as their optimization techniques, which hints the versatility of learned codecs. For practical utility, on-line model updating might be a critical factor for learned codecs to overcome the unpredictable domain shift between the distribution of training data and deployment environment.

\begin{figure}[!htb]
    \centering
	\subfigure[HEVC 0.065bpp 25.20dB]{
	\includegraphics[width=0.31\textwidth]{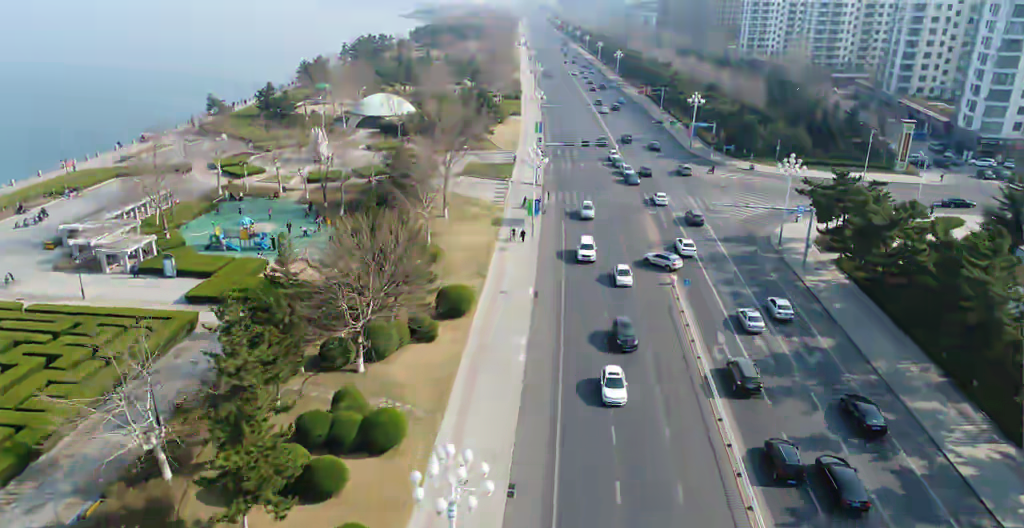}}
	\subfigure[OpenDVC 0.078bpp 25.99dB]{
	\includegraphics[width=0.31\textwidth]{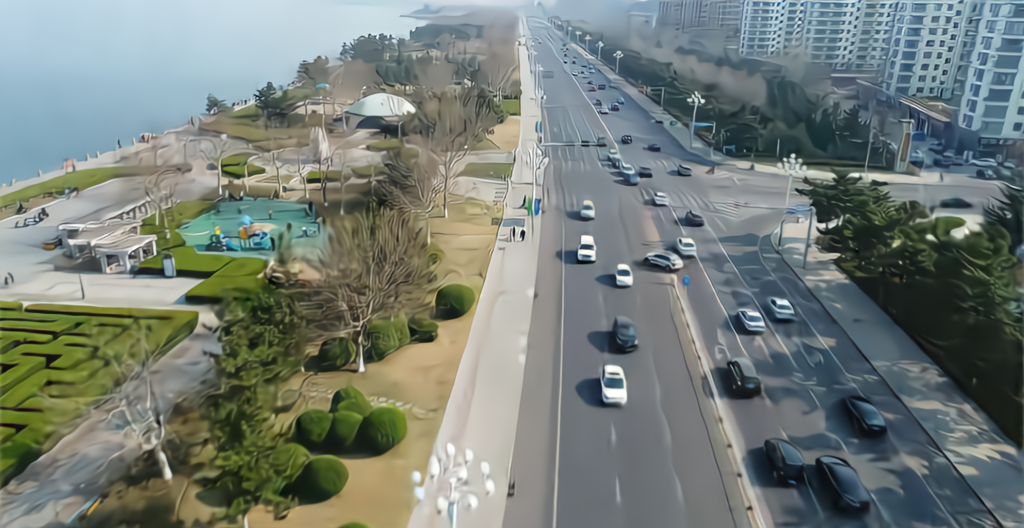}}
	\subfigure[EEV 0.077bpp 26.44dB]{
	\includegraphics[width=0.31\textwidth]{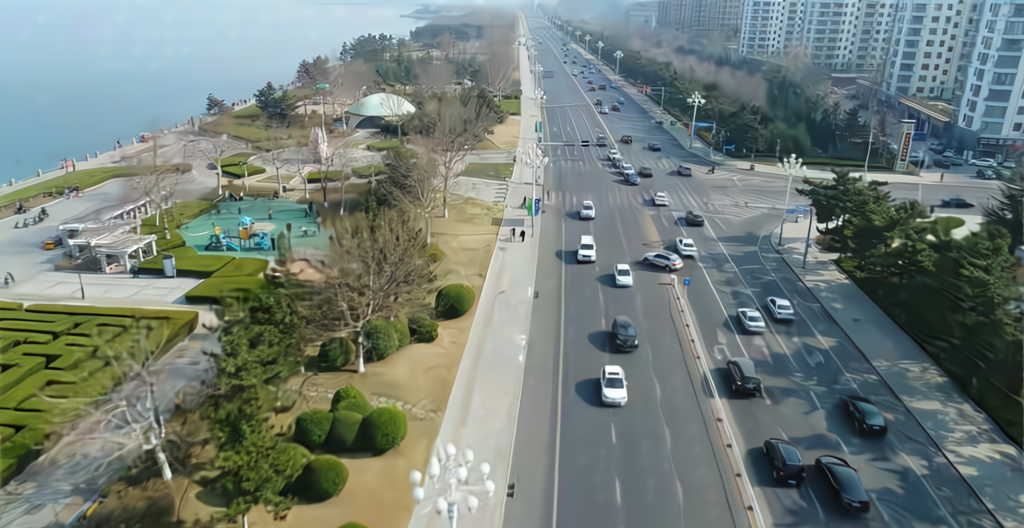}}    
	\subfigure[HEVC 0.055bpp 32.30dB]{
	\includegraphics[width=0.31\textwidth]{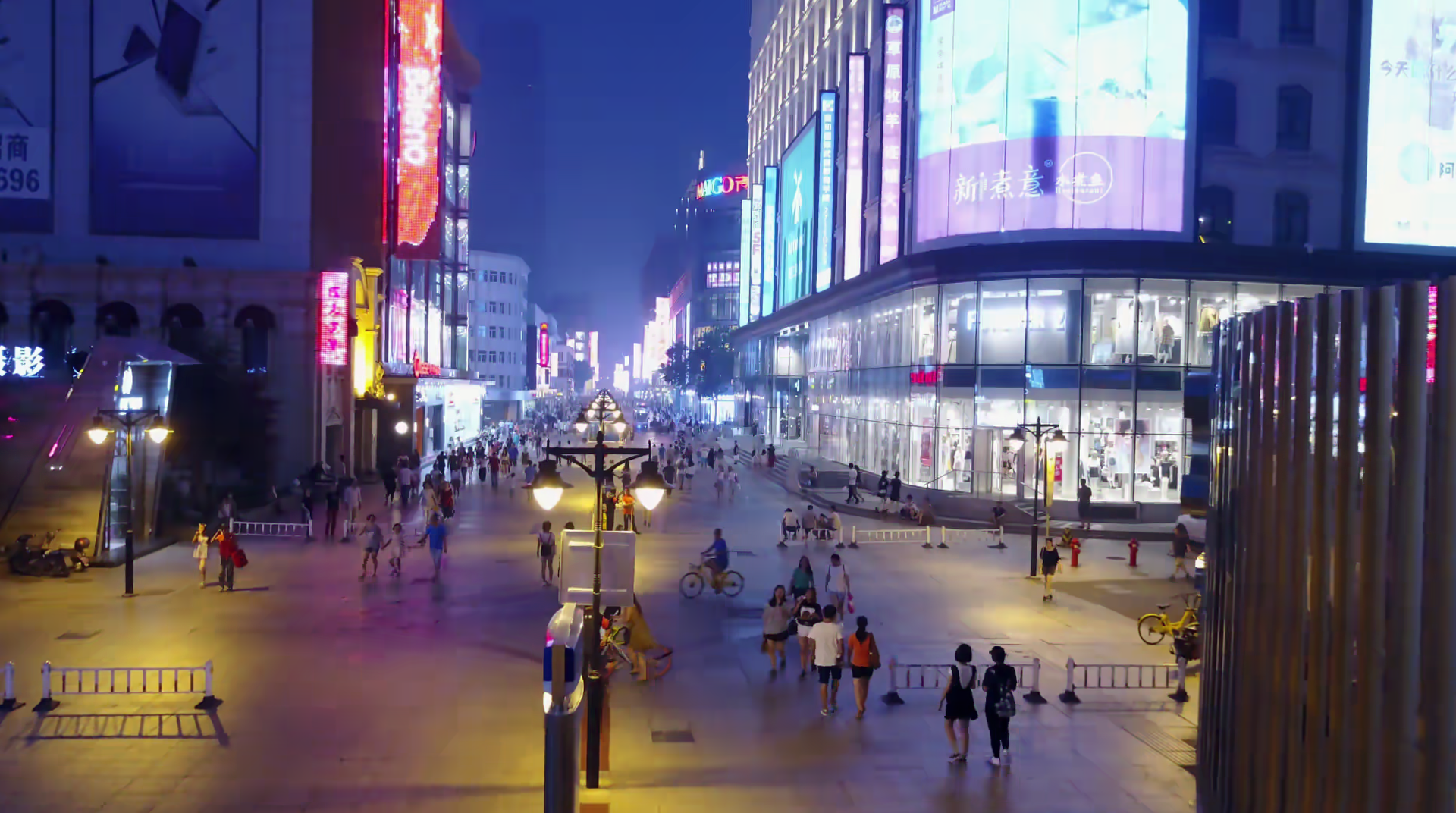}}
	\subfigure[OpenDVC 0.063bpp 32.84dB]{
	\includegraphics[width=0.31\textwidth]{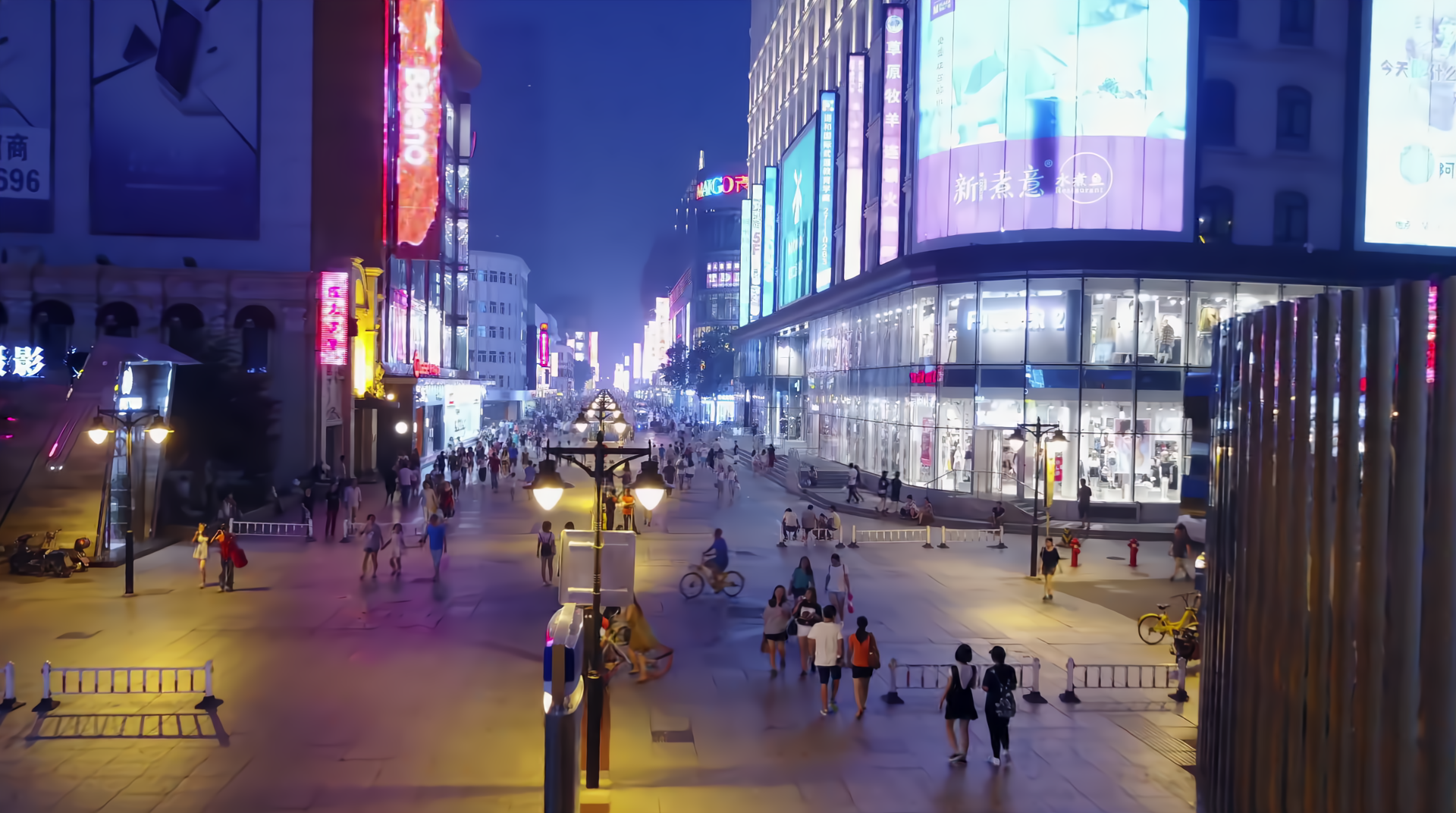}}
	\subfigure[EEV 0.058bpp 33.27dB]{
	\includegraphics[width=0.31\textwidth]{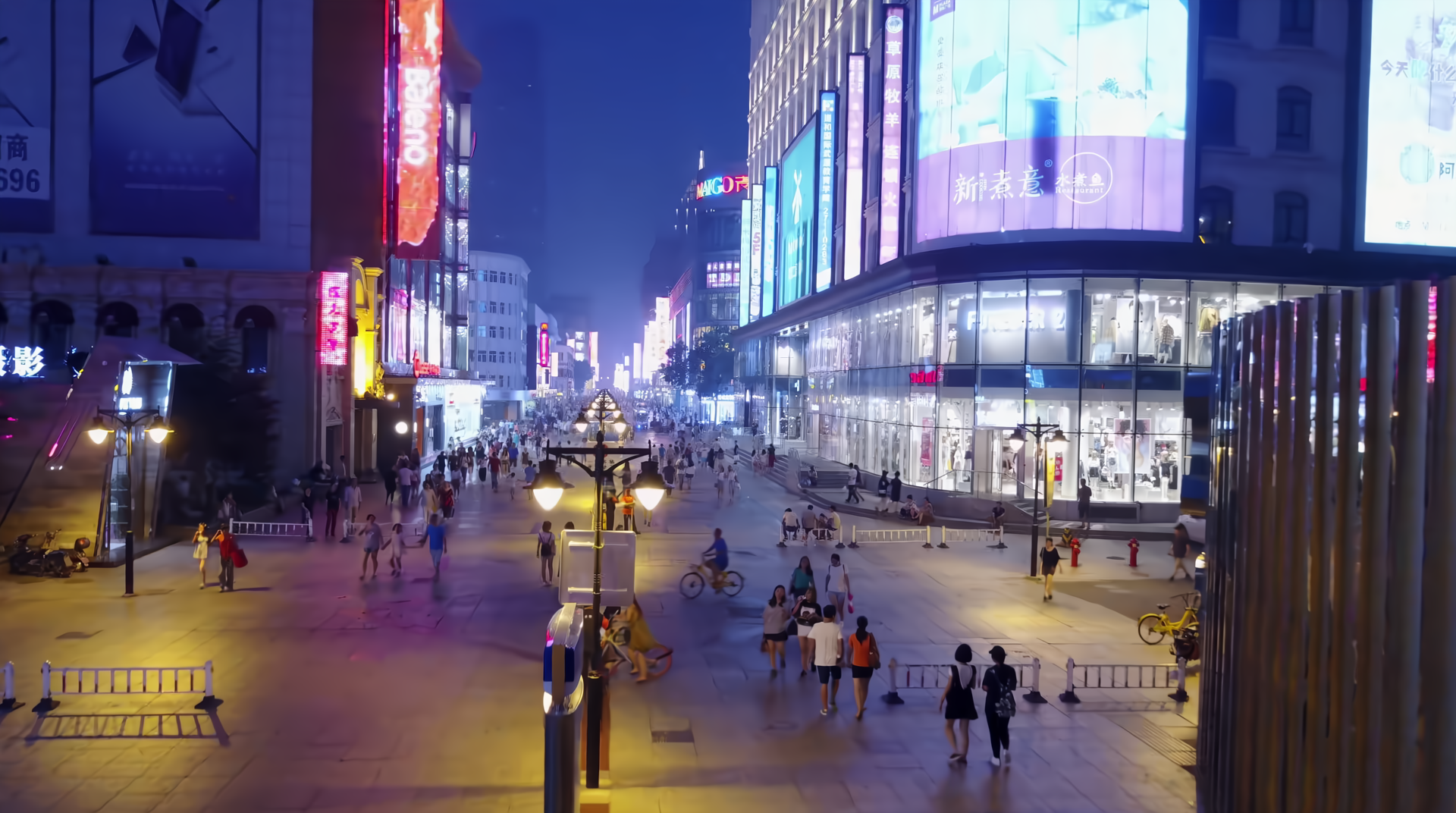}}
	
	\caption{Subjective quality comparison of reconstructed images of $RoadByTheSea$ and $NightMall$ for different codecs (HM-16.20-SCM-8.8~\cite{xu2015overview}, OpenDVC~\cite{yang2020opendvc}, and EEV). Zoom-in for better visualization.}
	\label{fig:visual-comparison}
\end{figure}

\SubSection{Visual Quality Evaluation}
The subjective quality of reconstructed images of $RoadByTheSea$ and $NightMall$ sequences for different codecs (HM-16.20-SCM-8.8~\cite{xu2015overview}, OpenDVC~\cite{yang2020opendvc}, and EEV) are shown in Fig.~\ref{fig:visual-comparison}. The first row represents the reconstructed frames of the 61$^{st}$ frame of $RoadByTheSea$ sequence. The test parameter setting for this set of visual comparison (Fig.~\ref{fig:visual-comparison}(a)-(c)) is QP=42 in HEVC-SCC, $\lambda$=256 for the two learned codecs. It is observed that the subjective quality of EEV codec is much better than the remaining two codecs, which shows the superiority of the learned codec. Specifically, textures of the EEV-coded picture are more visual pleasant, in which the road-lanes and other salient objects remain crystal clear while the conventional codec fails to restore such information. The blurring artifact also degrades the visual quality HEVC-SCC-coded picture. In this regard, the learned codec shows promising potentials for the autonomous drone video analysis or other downstream drone-vision tasks. The second row depicts the decoded pictures of the 23$^{nd}$ frame of $NightMall$ sequence. The test parameter setting for this set of comparison (Fig.~\ref{fig:visual-comparison}(d)-(f)) is QP=38 in HEVC-SCC, $\lambda$=512 for the two learned codecs. Surprisingly, the learned codec EEV obtains around 1dB PSNR improvement over HEVC-SCC when using similar bit-rate cost. The textures of pedestrians and bicycles can be kept after EEV compression. Given the subjective quality comparisons of all images in Fig.~\ref{fig:visual-comparison}, one can further conclude that the learned codecs are able to handle the UAV videos under different flying conditions (height, day/night time) even without fine-tuning on related datasets.

\Section{Working Mechanism of MPAI-EEV}
This work was accomplished in the MPAI-EEV coding project, which is an MPAI standard project seeking to compress video by exploiting AI-based data coding technologies1. Within this workgroup, experts around the globe gather and review the progress, and plan new efforts every two weeks. In its current phase, attendance at MPAI-EEV meetings is open to interested experts\footnote{http://eev.mpai.community}. Since its formal establishment in Nov. 2021, the MPAI EEV has released three major versions of it reference models.  MPAI-EEV plans on being an asset for AI-based end-to-end video coding by continuing to contribute new development in the end-to-end video coding field.

\Section{Conclusion}
In this paper, we build the first benchmark for the task of learning based UAV video coding, which consists of diverse UAV video contents associated with the R-D characteristics of classical and emerging learning based video codecs. It is observed that learned codecs outperform the conventional hybrid codecs for drone video compression since learned codecs have better model capacity in capturing the motion blur recorded by the drone-mounted cameras. The proposed benchmark has constructed a solid baseline for compressing UAV videos and facilitates the future research works for related topics such as downstream drone video analysis tasks.

\Section{References}
\bibliographystyle{IEEEbib}\tiny
\bibliography{refs}\tiny

\end{document}